\def\bhline{\specialrule{.2em}{0em}{0em}}

\documentclass[11pt,a4paper]{article}
\usepackage[hyperref]{acl2019}
\usepackage{times}
\usepackage{latexsym}
\usepackage{amssymb}
\usepackage{amsmath}
\usepackage{mathtools}
\usepackage{amsmath,amssymb}

\usepackage{booktabs}
\usepackage{caption}    
\usepackage{subcaption}    
\usepackage{graphicx}
\usepackage{appendix}
\usepackage{comment}
\usepackage[export]{adjustbox}

\usepackage{url}

\aclfinalcopy 
\def\aclpaperid{***} 

\newcommand{\inv}{\vspace{-0em}}

\title{Learning Bilingual Word Embeddings Using Lexical Definitions}

\author{Weijia Shi$^1$, Muhao Chen$^1$, Yingtao Tian$^2$, Kai-Wei Chang$^1$\\
$^1$Department of Computer Science, University of California, Los Angeles\\
$^2$Department of Computer Science, Stony Brook University\\
\{swj0419, muhaochen, kw2c\}@cs.ucla.edu; \{yittian\}@cs.stonybrook.edu\\
}
  
\def\model{BilLex (\textbf{Bil}ingual Word Embeddings Based on \textbf{Lex}ical Definitions)}

\def\modelname{BilLex}

\date{}

\begin{document}
\maketitle

\begin{abstract}
Bilingual word embeddings, which represent lexicons of different languages in a shared embedding space, are essential for supporting semantic and knowledge transfers in a variety of cross-lingual NLP tasks. Existing approaches to training bilingual word embeddings require often require pre-defined seed lexicons that are expensive to obtain, or parallel sentences that comprise coarse and noisy alignment. 
In contrast, we propose \modelname\ that leverages publicly available lexical definitions for bilingual word embedding learning.
Without the need of predefined seed lexicons, \modelname\ comprises a novel word pairing strategy to automatically identify and propagate the precise fine-grained word alignment from lexical definitions.
We evaluate \modelname\ in word-level and sentence-level translation tasks, which seek to find the cross-lingual counterparts of words and sentences respectively.
\modelname\ significantly outperforms previous embedding methods on both tasks.
\end{abstract}


\section{Introduction}

Bilingual word embeddings are the essential components of multilingual NLP systems.
These embeddings capture cross-lingual semantic transfers of words and phrases from bilingual corpora,
and are widely deployed in many NLP tasks, such as machine translation \cite{conneau2017word}, 
cross-lingual Wikification \cite{tsai2016cross}, knowledge alignment \cite{chen2018cotrain} and semantic search \cite{vulic2015monolingual}.
\par

A variety of approaches have been proposed to learn bilingual word embeddings 
\cite{duong2017multilingual,luong2015bilingual,coulmance2016trans}.
Many such approaches rely on the use of aligned corpora.
Such corpora could be seed lexicons that provide word-level mappings between two languages \cite{mikolov2013exploiting, xing2015normalized}, or parallel corpora that align sentences and documents \cite{Alexandre, Bilbowa}.
However, these methods critically suffer from several deficiencies.
First, seed-lexicon-based approaches 
are often hindered by the limitedness of seeds \cite{vulic2016role}, which is an intractable barrier since high-quality seed lexicons require extensive human efforts to obtain \cite{zhang2017bilingual}. 
Second, parallel corpora provide 
coarse alignment that does not often accurately infer fine-grained semantic transfers of lexicons \cite{ruder2017survey}. 



\begin{figure}
\centering
\begin{subfigure}[b]{.8\linewidth}
\includegraphics[width=1.1\linewidth,center]{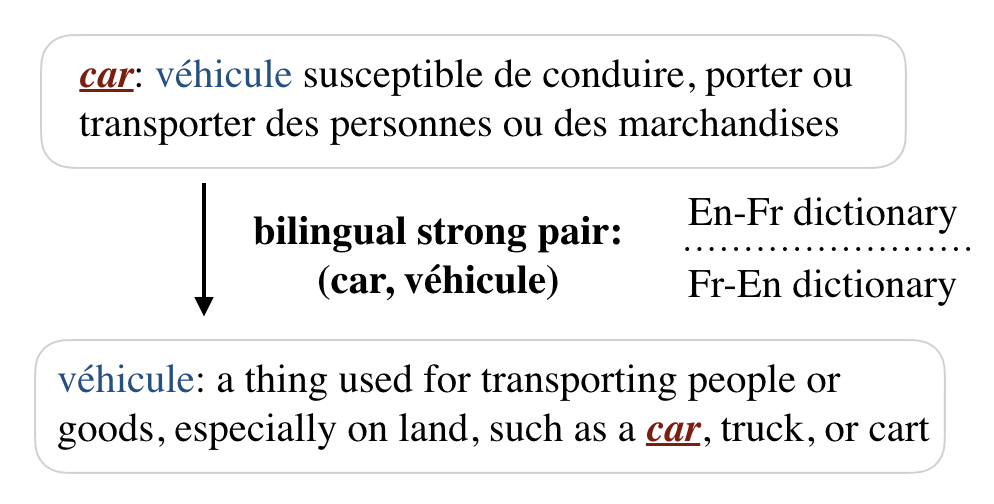}
\caption{Bilingual strong pair}
\end{subfigure} 
\begin{subfigure}[b]{.45\linewidth}
\includegraphics[width=1.1\linewidth,right]{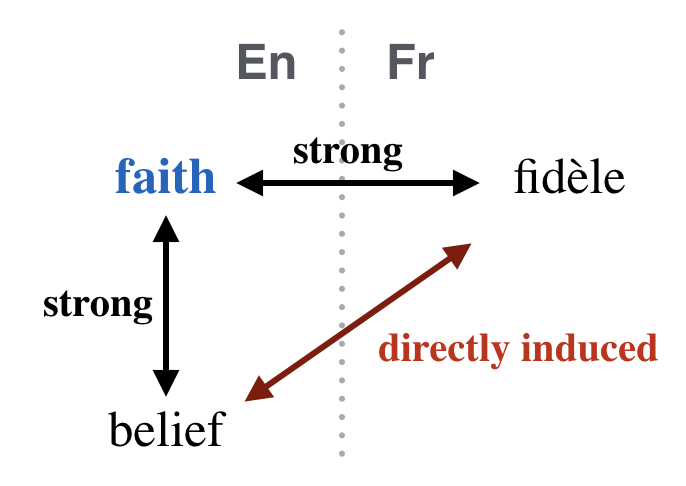}
\caption{Directly induced pair}
\end{subfigure} 
\begin{subfigure}[b]{.45\linewidth}
\includegraphics[width =1.1\linewidth,left]{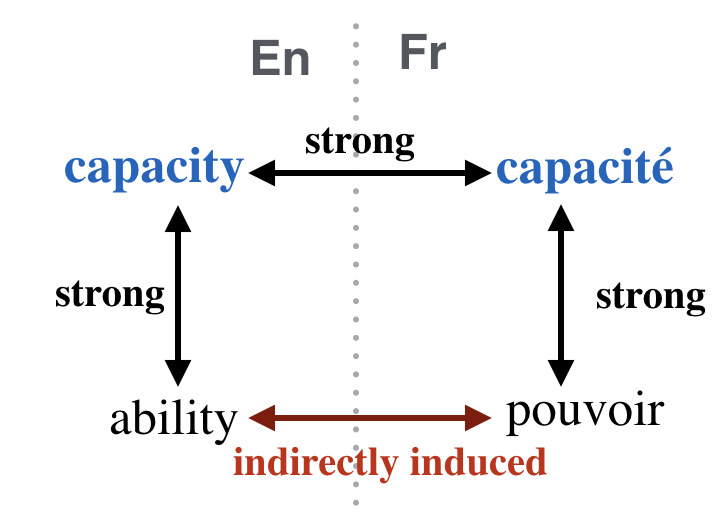}
\caption{Indirectly induced pair}
\end{subfigure} 
\caption{Examples of three types of word pairs. The blue words in (b-c) are pivot words of the induced pairs.} \label{fig:pairs}
\end{figure}

Unlike the existing methods, 
we propose to use publicly available dictionaries\footnote{We refer to \emph{dictionary} in its regular meaning, i.e. the collections of word definitions. This is different from some papers that refer to dictionaries as seed lexicons.} for bilingual word embedding learning.
Dictionaries, such as Wiktionary and Merriam-Webster,  
contain large collections of lexical definitions, which are clean linguistic knowledge that naturally connects word semantics within and across human languages.
Hence, dictionaries provide valuable information to bridge the lexicons in different languages.  
However, cross-lingual learning from lexical definitions is a non-trivial task.
A straightforward approach based on aligning the target word embedding
to the
aggregated embedding of words in the definition might work,
but not all words in a definition are semantically related to the defined target word (Fig.~\ref{fig:pairs}(a)).
Therefore, a successful model has to effectively identify the 
most related lexicons from the multi-granular and asymmetric alignment of lexical definitions.
Besides, how to leverage both bilingual and monolingual dictionaries for cross-lingual learning is another challenge.
\par

In this paper, we propose \model\ to learn bilingual word embeddings. 
\modelname\ constitutes a 
carefully designed two-stage mechanism to automatically cultivate, propagate and leverage lexicon pairs of high semantic similarity from lexical definitions in dictionaries.
It first extracts \emph{bilingual strong word pairs} from bilingual lexical definitions of which the words contribute to the cross-lingual definitions of each other. 
On top of that, our model automatically exploits 
\emph{induced word pairs}, which utilize
monolingual dictionaries and the aforementioned strong pairs to exploit semantically related word pairs.
This automated word pair induction process enables \modelname\ to capture abundant high-quality lexical alignment information,
based on which the cross-lingual semantic transfer of words is easily captured in a shared embedding space. Experimental results on word-level and sentence-level translation tasks show 
that \modelname\ drastically outperforms various baselines that are trained on parallel or seed-lexicon corpora,
as well as state-of-the-art unsupervised methods.

\section{Related Work}
Prior approaches to learning bilingual word embeddings often rely on
word or sentence alignment~\cite{ruder2017survey}. 
In particular, seed lexicon methods~\cite{mikolov2013exploiting,faruqui2014improving, guo2015cross} learn transformations across different language-specific embedding spaces based on predefined word alignment. 
The performance of these approaches is limited by the sufficiency of seed lexicons. Besides, parallel corpora
methods \cite{Bilbowa,coulmance2016trans} leverage the aligned sentences in different languages and force the representations of corresponding sentence components to be similar. 
However, 
aligned sentences merely provide weak alignment of lexicons that do not accurately capture the one-to-one mapping of words, while such a mapping is well-desired by translation tasks \cite{upadhyay2016cross}. 
In addition, a few unsupervised approaches 
alleviate the 
use of bilingual resources 
\cite{chen2018unsupervised,conneau2017word}. 
These models require considerable effort to train and rely heavily on massive monolingual corpora.

Monolingual lexical definitions have been 
used for weak supervision of monolingual word similarities \cite{tissier2017dict2vec}. Our work demonstrates that dictionary information can be extended to 
a cross-lingual scenario,
for which we develop a simple yet effective induction method to populate fine-grain word alignment.

\section{Modeling}
We first provide the formal definition of bilingual dictionaries. Let $\mathcal{L} $ be the set of languages and $\mathcal{L}^2$ be the set of ordered language pairs. For a language $l \in \mathcal{L}$, we use $V_l$ to denote its vocabulary, where for each word $w \in V_l$, $\textbf{w} \in \mathbb{R}^k$ denotes its embedding vector. A dictionary 
denoted as $D_{l_i,l_j}$ 
contains words in language $l_i$ and their 
definitions in $l_j$. In particular, $D_{l_i,l_j}$ is a monolingual dictionary if $l_i = l_j$ and is a bilingual dictionary if $l_i \neq l_j$. A dictionary $D_{l_i,l_j}$ contains dictionary entries $(w^i, Q^j(w^i))$, where $w^i \in V_{l_i}$ is the word being defined and $Q^j(w^i)$ is a sequence of words in $l_j$ describing the meaning of the word $w^i$.
Fig.~\ref{fig:pairs}(a) shows an entry from an English-French dictionary, and one from a French-English dictionary. 

\modelname\ allows us to exploit semantically related word pairs 
in two stages. 
We first use bilingual dictionaries to construct bilingual strong pairs, which are similar to those monolingual word pairs in \cite{tissier2017dict2vec}.
Then based on the given strong word pairs and monolingual dictionaries, we provide two types of induced word pairs to further enhance the cross-lingual learning.

\subsection{Bilingual Strong Pairs}
A bilingual strong pair contains two words with high semantic relevance. 
Such a pair of words that mutually contribute to the cross-lingual definitions of each other is defined as below.

\noindent \textbf{Definition (Bilingual Strong Pairs)} \textit{$P^S_{l_i,l_j}$ is the set of bilingual strong pairs in $(l_i, l_j) \in \mathcal{L}^2$ $(l_i \neq l_j)$, where each word pair is defined as: }
$(w^i, w^j) \in P^S_{l_i,l_j} \Leftrightarrow w^i \in Q^i(w^j) \wedge w^j \in Q^j(w^i)$

Intuitively, if $w^i$ appears in the cross-lingual definition of $w^j$ and $w^j$ appears in the cross-lingual definition of $w^i$, 
then $w^i$ and $w^j$ should be semantically close to each other.
Particularly, $P^S_{l_i,l_j}$ denotes monolingual strong pairs if $l_i = l_j$. 
For instance, \textit{(car, v{\'e}hicule)} depicted in Fig.~\ref{fig:pairs}(a) form a bilingual strong pair. 
Note that \citeauthor{tissier2017dict2vec} also introduce the monolingual weak pairs by pairing the target word with the other words from its definition, which do not form strong pairs with it. 
However, 
we do not extend such weak pairs to the bilingual setting, as we find them to be inaccurate to represent cross-lingual corresponding words.



\subsection{Induced Word Pairs}
Since bilingual lexical definitions cover only limited numbers of words in two languages,
we 
incorporate both monolingual and bilingual strong pairs, from which we induce two types of word pairs with different confidence: \textit{directly induced pairs} and \textit{indirectly induced pairs}.  




\noindent \textbf{Definition (Bilingual Directly Induced Pairs)}
\textit{$P^D_{l_i,l_j}$ is the set of bilingual directly induced pairs in $(l_i, l_j) \in \mathcal{L}^2$, where each word pair is defined as: }
$(w^i, w^j) \in P^D_{l_i,l_j} \Leftrightarrow \exists w^i_p, (w^i, w^i_p) \in P^S_{l_i,l_i}   \wedge (w^i_p, w^j) \in P^S_{l_i,l_j}$

Intuitively, a bilingual induced pair $(w^i, w^j)$ indicates that
we can find a pivot word that forms a monolingual strong pair with one word from $(w^i, w^j)$ and a bilingual strong pair with the other. 

\noindent \textbf{Definition (Bilingual Indirectly Induced Pairs)}
\textit{$P^I_{l_i,l_j}$ is the set of bilingual indirectly induced pairs in $(l_i, l_j) \in \mathcal{L}^2$, where each word pair is defined as: }
$(w^i, w^j) \in P^I_{l_i,l_j} \Leftrightarrow \exists (w^i_p, w^j_p) \in P^S_{l_i,l_j},
(w^i, w^i_p) \in P^S_{l_i,l_i} \wedge (w^j, w^j_p) \in P^S_{l_j,l_j}
$

A bilingual indirectly induced pair $(w^i, w^j)$ indicates that there exists a pivot bilingual strong pair $(w^i_p, w^j_p)$, such that $w^i_p$ forms a monolingual strong pair with $w_i$ and $w^j_p$ forms a monolingual strong pair with $w^j$. 
Fig.~\ref{fig:pairs}(b-c) shows examples of the two types of induced word pairs.


\subsection{Training}
Our model jointly learns three word-pair-based cross-lingual objectives $\Omega_{K}$ to align the embedding spaces of two languages, and two monolingual monolingual Skip-Gram losses \cite{mikolov2013distributed} $L_{l_i}, L_{l_j}$ to 
preserve monolingual word similarities.
Given a language pair $(l_i, l_j) \in \mathcal{L}^2$, the learning objective of \modelname\ is to minimize the following joint loss function:

\inv
\begin{equation*}
J = L_{l_i} + L_{l_j} + \sum_{K\in\{P^S,P^D,P^I\}} \lambda_K \Omega_{K}
\end{equation*}
\inv

Each $\lambda_K$ ($K\in\{P^S,P^D,P^I\}$) thereof, is the hyperparameter that controls how much the corresponding type of word pairs contributes to cross-lingual learning. 
For alignment objectives, we use word pairs in both directions of an ordered language pair $(l_i, l_j) \in \mathcal{L}^2$ to capture the cross-lingual semantic similarity of words, such that $P^S = P^S_{l_i, l_j} \cup  P^S_{l_j, l_i}, P^D = P^D_{l_i, l_j} \cup  P^D_{l_j, l_i}$ and $P^I = P^I_{l_i, l_j} \cup  P^I_{l_j, l_i}$.
Then for each $K\in\{P^S,P^D,P^I\}$, the alignment objective $\Omega_K$ is defined as below, where $\sigma$ is the sigmoid function.

\inv
\begin{align*}
\Omega_K &= - \frac{1}{|K|} \sum_{(w^i,w^j) \in K} \Big( \log  \sigma({\textbf{w}^{i}}^\top  \textbf{w}^j)  \\
&+ \sum_{ (w_a, w_b) \in N_i(w^j) \cup N_j(w^i)} \log\sigma({-\textbf{w}_a}^\top \textbf{w}_b) 
 \Big) 
\end{align*}
\inv

For each word pair $(w^i, w^j)$, we use the unigram distribution raised to the power of 0.75 to select a number of words in $l_j$ (or $l_i$) for $w^i$ (or $w^j$) to form a negative sample set $N_i(w^j)$ (or $N_j(w^i)$). 
Without loss of generality, we define the negative sample set as $N_i(w^j) = \{ (w^i_n, w^j) | w^i_n \sim  \mathrm{U}_i(w) \wedge (w^i_n, w^j) \notin P^S \cup P^D \cup P^I \}$, where $\mathrm{U}_i(w)$ is the distribution of words in $l_i$.



\section{Experiment}
We evaluate \modelname\ on two bilingual tasks: \emph{word translation} and \emph{sentence translation retrieval}. 
Following the convention \cite{Bilbowa,mogadala2016bilingual}, we evaluate \modelname\ between English-French and English-Spanish.
Accordingly, we extract word pairs from both directions of bilingual dictionaries in Wiktionary for these language pairs.
To support the induced word pairs,
we also extract monolingual lexical definitions in the three languages involved, which include 238k entries in English, 107k entries in French and 49k entries in Spanish. 
The word pair extraction process of \modelname\ excludes stop words and punctuation in the lexical definitions.
The statistics of three types of extracted word pairs are reported in Table~\ref{tbl:statistics}.

\subsection{Word translation}
This task aims to retrieve the translation of a source word in the target language. 
We use the test set provided by \citeauthor{conneau2017word} \shortcite{conneau2017word}, which selects the most frequent 200k words of each language as candidates for 1.5k query words. 
We translate a query word by retrieving its k nearest neighbours in the target language, and report $P@k$ $(k = 1, 5)$ to represent the fraction of correct translations that are ranked not larger than $k$. 

{
\begin{table}
\setlength\tabcolsep{2pt}
\centering
\scriptsize
\begin{tabular}{c|c|ccc}
\bhline
Lang&\#Def&$S$&$I_D$&$I_I$\\
\hline
En\&Fr&108,638&52,406&48,524&62,488\\
En\&Es&56,239&32,210&29,857&37,952\\
\bhline
\end{tabular}
\vspace{-0.5em}
\caption{Statistics of dictionaries and word pair sets.}\label{tbl:statistics}
\vspace{-1em}
\end{table}
}


{
\begin{table}[t]
\setlength\tabcolsep{1pt}
\centering
\scriptsize
\begin{tabular}{c|cc|cc|cc|cc}
\bhline
Language&\multicolumn{2}{c|}{En-Fr}&\multicolumn{2}{c|}{Fr-En}&\multicolumn{2}{c|}{En-Es}&\multicolumn{2}{c}{Es-En}\\
\hline
Metric&$P@1$&$P@5$&$P@1$&$P@5$&$P@1$&$P@5$&$P@1$&$P@5$\\
\bhline
BiCVM&41.8&56.6&42.6&58.4&37.8&52.8&39.9&54.2\\
BilBOWA&42.3&59.7&45.2&59.2&37.6&50.3&45.8&53.7\\
BiSkip&44.0&58.4&45.9&60.2&41.2&58.0&45.4&56.9\\
Supervised MUSE&74.9&89.8&76.1&90.9&77.4&93.8&77.3&93.6\\
Unsupervised MUSE &\textbf{78.1}&\textbf{94.3}&\textbf{78.2}&\textbf{93.0}&\textbf{81.7}&\textbf{94.4}&\textbf{83.3}&\textbf{96.6}\\
\hline
\modelname($P^S$) &62.4&79.2&61.8&77.4&64.3&78.4&61.9&78.0\\
\modelname($P^S$+$P^D$) &73.6&87.3&75.3&87.7&73.7&88.7&76.0&90.2\\
\modelname($P^S$+$P^D$+$P^I$) &\textbf{82.5}&\textbf{96.2}&\textbf{83.8}&\textbf{96.0}&\textbf{82.0}&\textbf{96.5}&\textbf{85.1}&\textbf{96.8}\\
\bhline
\end{tabular}
\vspace{-0.5em}
\caption{Results of the word translation task.}\label{tbl:word}
\vspace{-1em}
\end{table}
}

{
\begin{table}[t]
\setlength\tabcolsep{1pt}
\centering
\scriptsize
\begin{tabular}{c|cc|cc}
\bhline
Language&\multicolumn{2}{c|}{En-Fr}&\multicolumn{2}{c}{Fr-En}\\
\hline
Metric&$P@1$&$P@5$&$P@1$&$P@5$\\
\bhline
BiCVM&24.4&40.5&32.3&43.8\\
BilBOWA&27.7&41.4&31.5&47.0\\
BiSkip&25.3&38.8&26.4&40.4\\
Supervised MUSE&\textbf{63.2}&\textbf{76.9}&\textbf{74.9}&85.4\\
Unsupervised MUSE&60.0&76.3&73.7&\textbf{87.6}\\
\hline
\modelname($P^S$) &47.4&59.7&57.2&69.6\\
\modelname($P^S$+$P^D$) &58.7&73.8&67.6&78.9\\
\modelname($P^S$+$P^D$+$P^I$) &\textbf{64.9}&\textbf{78.2}&\textbf{76.3}&\textbf{89.7}\\
\bhline
\end{tabular}
\vspace{-0.5em}
\caption{Results of sentence translation retrieval.}\label{tbl:sent}
\vspace{-1em}
\end{table}
}

\noindent \textbf{Evaluation protocol.} The hyperparameters of \modelname\ are tuned based on a small validation set of 1k word pairs provided by \citeauthor{conneau2017word} \shortcite{conneau2017word}. 
We allocate 128-dimensional word embeddings with pre-trained BilBOWA \cite{Bilbowa}.
and use the standard configuration to Skip-Gram   \cite{mikolov2013distributed} on monolingual Wikipedia dumps. 
We set the negative sampling size of bilingual word pairs to 4, which is selected from 0 to 10 with the step of 1.
$\lambda_{P^S}$ is set to $0.9$, which is tuned from 0 to 1 with the step of 0.1. 
As we assume that the strong pair relations between words are independent, we empirically set $\lambda_{P^D}$ = ${(\lambda_{P^S})}^2 = 0.81$ and $\lambda_{P^I}$ = ${(\lambda_{P^S})}^3 = 0.729$. 
We minimize the loss function using AMSGrad \cite{reddi2018convergence} with a learning rate of 0.001.
The training is terminated based on early stopping. 
We limit the vocabularies as the 200k most frequent words in each language, and exclude the bilingual strong pairs that have appeared in the test set.
The baselines we compare against include BiCVM \cite{hermann2014multilingual}, BilBOWA \cite{Bilbowa}, Biskip \cite{luong2015bilingual}, supervised and unsupervised MUSE \cite{conneau2017word}. 

\noindent \textbf{Results. }Results are summarized in Table~\ref{tbl:word}, where the performance of \modelname\ is reported for three variants: (i) training with bilingual strong pairs only \modelname($P^S$), (ii) 
with directly induced pair added \modelname($P^S$+$P^D$), and (iii)
with all three types of word pairs \modelname($P^S$+$P^D$+$P^I$).  \modelname($P^S$+$P^D$+$P^I$) thereof, offers consistently better performance in all settings, which implies that the induced word pairs are effective in improving the cross-lingual learning of lexical semantics. 
Among the baseline models, the unsupervised MUSE outperforms the other four supervised ones. We also discover that for the word translation task, the supervised models with coarse alignment such as BiCVM and BilBOWA do not perform as well as the models with word-level supervision, such as BiSkip and supervised MUSE. Our best \modelname\ outperforms unsupervised MUSE by 4.4$\sim$5.7\% of $P@1$ between En and Fr, and by 0.3$\sim$1.8\% between En and Es.
The reason why the settings between En and Fr achieve better performance is that there are much fewer bilingual definitions between En and Es.

\subsection{Sentence translation retrieval}
This task focuses on retrieving the sentence in the target language space with the tf-idf weighted sentence representation approach. We follow the experiment setup in \cite{conneau2017word} with 2k source sentence queries and 200k target sentences from the Europarl corpus for English and French.
We carry forward model 
configurations 
from the previous experiment, and report $P@k$ $(k = 1, 5)$.

\noindent \textbf{Results. }The results 
are reported in Table~\ref{tbl:sent}. 
Overall, our best model variant \modelname($P^S$+$P^D$+$P^I$) performs better than the best baseline with a noticeable increment of $P@1$ by 1.4$\sim$1.7\% and P@5 by 1.3$\sim$2.1\%. 
This demonstrates that \modelname\ is suitable for transferring 
sentential semantics.




\section{Conclusion}
In this paper, we propose \modelname, 
a novel bilingual word embedding model 
based on lexical definitions. \modelname\ is motivated by the fact that openly available dictionaries offer high-quality linguistic knowledge to connect lexicons across languages. We 
design the word pair induction method to capture semantically related lexicons in dictionaries, 
which serve as alignment information
in joint training. \modelname\ outperforms state-of-the-art 
methods on word and sentence translation tasks. 
\bibliography{ref}

\begin{thebibliography}{23}
\expandafter\ifx\csname natexlab\endcsname\relax\def\natexlab#1{#1}\fi

\bibitem[{Chen et~al.(2018)Chen, Tian, Chang, Skiena et~al.}]{chen2018cotrain}
Muhao Chen, Yingtao Tian, Kai-Wei Chang, Steven Skiena, et~al. 2018.
\newblock Co-training embeddings of knowledge graphs and entity descriptions
  for cross-lingual entity alignment.
\newblock In \emph{Proceedings of the 27th International Joint Conference on
  Artificial Intelligence}, pages 3998--4004.

\bibitem[{Chen and Cardie(2018)}]{chen2018unsupervised}
Xilun Chen and Claire Cardie. 2018.
\newblock Unsupervised multilingual word embeddings.
\newblock In \emph{Proceedings of the 2018 Conference on Empirical Methods in
  Natural Language Processing}, pages 261--270.

\bibitem[{Conneau et~al.(2018)Conneau, Lample, Ranzato, Denoyer, and
  J{\'e}gou}]{conneau2017word}
Alexis Conneau, Guillaume Lample, Marc'Aurelio Ranzato, Ludovic Denoyer, and
  Herv{\'e} J{\'e}gou. 2018.
\newblock Word translation without parallel data.
\newblock \emph{ICLR}.

\bibitem[{Coulmance et~al.(2015)Coulmance, Marty, Wenzek, and
  Benhalloum}]{coulmance2016trans}
Jocelyn Coulmance, Jean-Marc Marty, Guillaume Wenzek, and Amine Benhalloum.
  2015.
\newblock Trans-gram, fast cross-lingual word-embeddings.

\bibitem[{Duong et~al.(2017)Duong, Kanayama, Ma, Bird, and
  Cohn}]{duong2017multilingual}
Long Duong, Hiroshi Kanayama, Tengfei Ma, Steven Bird, and Trevor Cohn. 2017.
\newblock Multilingual training of crosslingual word embeddings.
\newblock In \emph{Proceedings of the 15th Conference of the European Chapter
  of the Association for Computational Linguistics: Volume 1, Long Papers},
  volume~1, pages 894--904.

\bibitem[{Faruqui and Dyer(2014)}]{faruqui2014improving}
Manaal Faruqui and Chris Dyer. 2014.
\newblock Improving vector space word representations using multilingual
  correlation.
\newblock In \emph{Proceedings of the 14th Conference of the European Chapter
  of the Association for Computational Linguistics}, pages 462--471.

\bibitem[{Gouws et~al.(2015)Gouws, Bengio, and Corrado}]{Bilbowa}
Stephan Gouws, Yoshua Bengio, and Greg Corrado. 2015.
\newblock Bilbowa: Fast bilingual distributed representations without word
  alignments.
\newblock In \emph{Inter National Conference on Machine Learning}.

\bibitem[{Guo et~al.(2015)Guo, Che, Yarowsky, Wang, and Liu}]{guo2015cross}
Jiang Guo, Wanxiang Che, David Yarowsky, Haifeng Wang, and Ting Liu. 2015.
\newblock Cross-lingual dependency parsing based on distributed
  representations.
\newblock In \emph{Proceedings of the 53rd Annual Meeting of the Association
  for Computational Linguistics and the 7th International Joint Conference on
  Natural Language Processing (Volume 1: Long Papers)}, volume~1, pages
  1234--1244.

\bibitem[{Hermann and Blunsom(2014)}]{hermann2014multilingual}
Karl~Moritz Hermann and Phil Blunsom. 2014.
\newblock Multilingual models for compositional distributed semantics.
\newblock In \emph{Proceedings of the 52nd Annual Meeting of the Association
  for Computational Linguistics (Volume 1: Long Papers)}, volume~1, pages
  58--68.

\bibitem[{Klementiev et~al.(2012)Klementiev, Titov, and Bhattarai}]{Alexandre}
Alexandre Klementiev, Ivan Titov, and Binod Bhattarai. 2012.
\newblock Inducing crosslingual distributed representations of words.
\newblock In \emph{Proceedings of COLING 2012}.

\bibitem[{Luong et~al.(2015)Luong, Pham, and Manning}]{luong2015bilingual}
Thang Luong, Hieu Pham, and Christopher~D Manning. 2015.
\newblock Bilingual word representations with monolingual quality in mind.
\newblock In \emph{Proceedings of NAACL-HLT}, pages 151--159.

\bibitem[{Mikolov et~al.(2013{\natexlab{a}})Mikolov, Le, and
  Sutskever}]{mikolov2013exploiting}
Tomas Mikolov, Quoc~V Le, and Ilya Sutskever. 2013{\natexlab{a}}.
\newblock Exploiting similarities among languages for machine translation.
\newblock \emph{CoRR,abs/1309.4168}.

\bibitem[{Mikolov et~al.(2013{\natexlab{b}})Mikolov, Sutskever, Chen, Corrado,
  and Dean}]{mikolov2013distributed}
Tomas Mikolov, Ilya Sutskever, Kai Chen, Greg~S Corrado, and Jeff Dean.
  2013{\natexlab{b}}.
\newblock Distributed representations of words and phrases and their
  compositionality.
\newblock In \emph{Advances in neural information processing systems}, pages
  3111--3119.

\bibitem[{Mogadala and Rettinger(2016)}]{mogadala2016bilingual}
Aditya Mogadala and Achim Rettinger. 2016.
\newblock Bilingual word embeddings from parallel and non-parallel corpora for
  cross-language text classification.
\newblock In \emph{Proceedings of the 2016 Conference of the North American
  Chapter of the Association for Computational Linguistics: Human Language
  Technologies}, pages 692--702.

\bibitem[{Reddi et~al.(2018)Reddi, Kale, and Kumar}]{reddi2018convergence}
Sashank~J Reddi, Satyen Kale, and Sanjiv Kumar. 2018.
\newblock On the convergence of adam and beyond.
\newblock In \emph{ICLR}.

\bibitem[{Ruder et~al.(2017)Ruder, Vuli{\'c}, and S{\o}gaard}]{ruder2017survey}
Sebastian Ruder, Ivan Vuli{\'c}, and Anders S{\o}gaard. 2017.
\newblock A survey of cross-lingual word embedding models.
\newblock \emph{Journal of Artificial Intelligence Research}.

\bibitem[{Tissier et~al.(2017)Tissier, Gravier, and
  Habrard}]{tissier2017dict2vec}
Julien Tissier, Christophe Gravier, and Amaury Habrard. 2017.
\newblock Dict2vec: Learning word embeddings using lexical dictionaries.
\newblock In \emph{Conference on Empirical Methods in Natural Language
  Processing (EMNLP 2017)}, pages 254--263.

\bibitem[{Tsai and Roth(2016)}]{tsai2016cross}
Chen-Tse Tsai and Dan Roth. 2016.
\newblock Cross-lingual wikification using multilingual embeddings.
\newblock In \emph{Proceedings of the 2016 Conference of the North American
  Chapter of the Association for Computational Linguistics: Human Language
  Technologies}, pages 589--598.

\bibitem[{Upadhyay et~al.(2016)Upadhyay, Faruqui, Dyer, and
  Roth}]{upadhyay2016cross}
Shyam Upadhyay, Manaal Faruqui, Chris Dyer, and Dan Roth. 2016.
\newblock Cross-lingual models of word embeddings: An empirical comparison.
\newblock In \emph{Proceedings of the 54th Annual Meeting of the Association
  for Computational Linguistics (Volume 1: Long Papers)}, volume~1, pages
  1661--1670.

\bibitem[{Vuli{\'c} and Korhonen(2016)}]{vulic2016role}
Ivan Vuli{\'c} and Anna Korhonen. 2016.
\newblock On the role of seed lexicons in learning bilingual word embeddings.
\newblock In \emph{Proceedings of the 54th Annual Meeting of the Association
  for Computational Linguistics (Volume 1: Long Papers)}, volume~1, pages
  247--257.

\bibitem[{Vuli{\'c} and Moens(2015)}]{vulic2015monolingual}
Ivan Vuli{\'c} and Marie-Francine Moens. 2015.
\newblock Monolingual and cross-lingual information retrieval models based on
  (bilingual) word embeddings.
\newblock In \emph{Proceedings of the 38th international ACM SIGIR conference
  on research and development in information retrieval}, pages 363--372. ACM.

\bibitem[{Xing et~al.(2015)Xing, Wang, Liu, and Lin}]{xing2015normalized}
Chao Xing, Dong Wang, Chao Liu, and Yiye Lin. 2015.
\newblock Normalized word embedding and orthogonal transform for bilingual word
  translation.
\newblock In \emph{Proceedings of the 2015 Conference of the North American
  Chapter of the Association for Computational Linguistics: Human Language
  Technologies}, pages 1006--1011.

\bibitem[{Zhang et~al.(2017)Zhang, Peng, Liu, Luan, and
  Sun}]{zhang2017bilingual}
Meng Zhang, Haoruo Peng, Yang Liu, Huan-Bo Luan, and Maosong Sun. 2017.
\newblock Bilingual lexicon induction from non-parallel data with minimal
  supervision.
\newblock In \emph{AAAI}, pages 3379--3385.

\end{thebibliography}
\bibliographystyle{acl_natbib}

\end{document}